\documentclass[11pt,a4paper]{article}
\usepackage{acl2015}
\usepackage{times}
\usepackage{url}
\usepackage{latexsym}
\usepackage{amsmath,amssymb}
\usepackage{caption}
\usepackage{courier}
\usepackage{graphicx}
\usepackage{epstopdf}
\usepackage{helvet}
\usepackage{url}
\usepackage{subcaption}
\usepackage{times}
\usepackage{multirow}
\usepackage{color}

\makeatletter
\newcommand{\xLeftrightarrow}[2][]{\ext@arrow 3359\Leftrightarrowfill@{#1}{#2}}
\newcommand{\xRightarrow}[2][]{\ext@arrow 3359\Rightarrowfill@{#1}{#2}}
\makeatother

\title{Modeling Relation Paths for Representation Learning of Knowledge Bases}
\author{Yankai Lin$^{1}$, Zhiyuan Liu$^{1}$ \thanks{\ \ Corresponding author: Z. Liu (liuzy@tsinghua.edu.cn)} , Huanbo Luan$^{1}$, Maosong Sun$^{1}$, Siwei Rao$^{2}$, Song Liu$^{2}$ \\
$^{1}$ {\small Department of Computer Science and Technology, State Key Lab on Intelligent Technology and Systems}, \\
{\small National Lab for Information Science and Technology, Tsinghua University, Beijing, China} \\
$^{2}$ {\small Samsung R\&D Institute of China, Beijing, China}
}

\begin{document}
\maketitle
\begin{abstract}
Representation learning of knowledge bases aims to embed both entities and relations into a low-dimensional space. Most existing methods only consider direct relations in representation learning. We argue that multiple-step relation paths also contain rich inference patterns between entities, and propose a path-based representation learning model. This model considers relation paths as translations between entities for representation learning, and addresses two key challenges: (1) Since not all relation paths are reliable, we design a path-constraint resource allocation algorithm to measure the reliability of relation paths. (2) We represent relation paths via semantic composition of relation embeddings. Experimental results on real-world datasets show that, as compared with baselines, our model achieves significant and consistent improvements on knowledge base completion and relation extraction from text. The source code of this paper can be obtained from \url{https://github.com/mrlyk423/relation_extraction}.
\end{abstract}

\section{Introduction}
People have recently built many large-scale knowledge bases (KBs) such as Freebase, DBpedia and YAGO. These KBs consist of facts about the real world, mostly in the form of triples, e.g., (\emph{Steve Jobs}, \texttt{FounderOf}, \emph{Apple Inc.}). KBs are important resources for many applications such as question answering and Web search. Although typical KBs are large in size, usually containing thousands of relation types, millions of entities and billions of facts (triples), they are far from complete. Hence, many efforts have been invested in relation extraction to enrich KBs.

Recent studies reveal that, neural-based representation learning methods are scalable and effective to encode relational knowledge with low-dimensional representations of both entities and relations, which can be further used to extract unknown relational facts. TransE \cite{bordes2013translating} is a typical method in the neural-based approach, which learns vectors (i.e., embeddings) for both entities and relations. The basic idea behind TransE is that, the relationship between two entities corresponds to a translation between the embeddings of the entities, that is, $\mathbf{h} + \mathbf{r} \approx \mathbf{t}$ when the triple (h, r, t) holds. Since TransE has issues when modeling 1-to-N, N-to-1 and N-to-N relations, various methods such as TransH \cite{wang2014knowledge} and TransR \cite{lin2015learning} are proposed to assign an entity with different representations when involved in various relations.

Despite their success in modeling relational facts, TransE and its extensions only take direct relations between entities into consideration. It is known that there are also substantial multiple-step relation paths between entities indicating their semantic relationships. The relation paths reflect complicated inference patterns among relations in KBs. For example, the relation path $h \xrightarrow{\texttt{BornInCity}} e_1 \xrightarrow{\texttt{CityInState}} e_2 \xrightarrow{\texttt{StateInCountry}} t$ indicates  the relation \texttt{Nationality} between $h$ and $t$, i.e., ($h$, \texttt{Nationality}, $t$).

\begin{figure}[htb]
\centering
\includegraphics[width=\columnwidth]{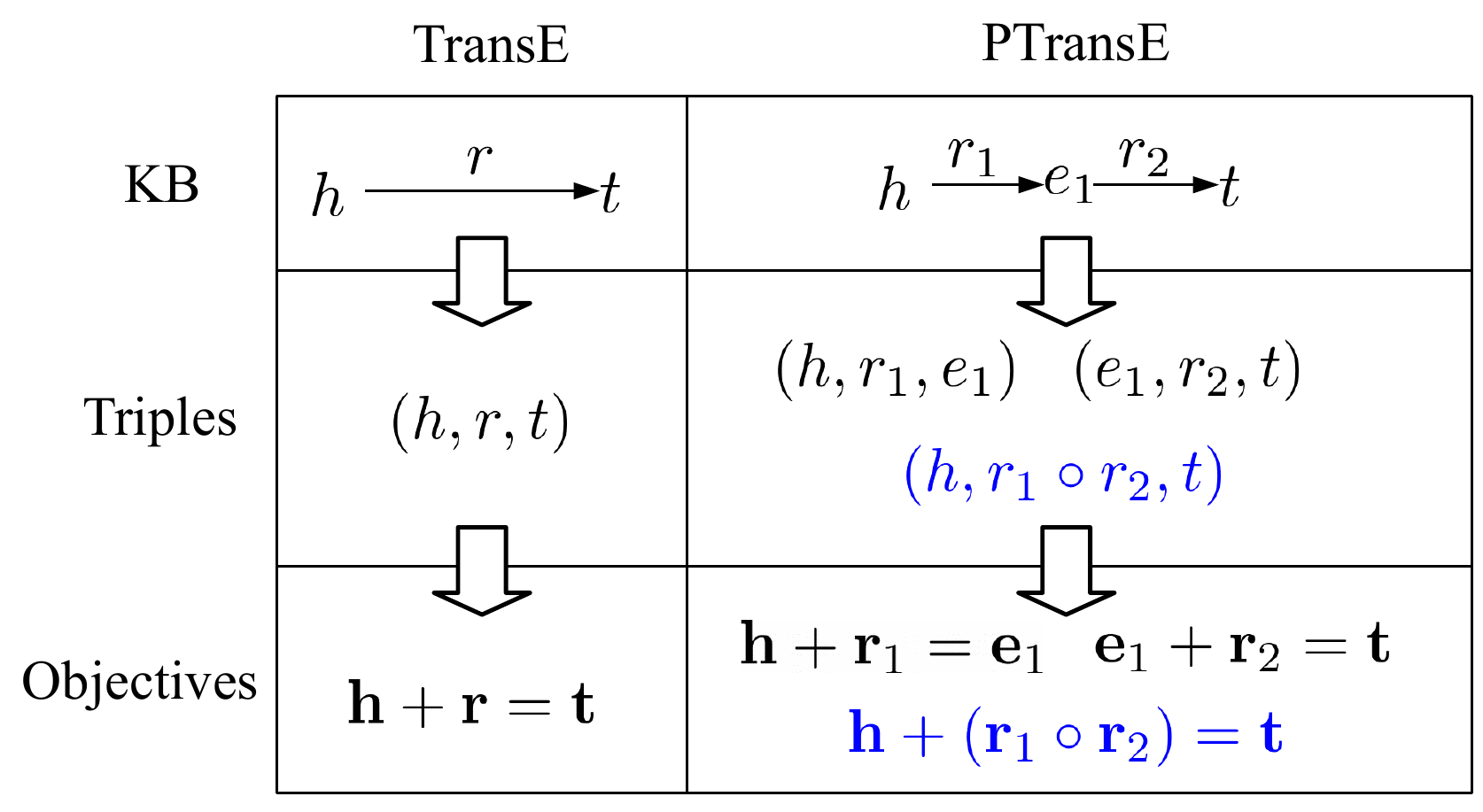}
\caption{TransE and PTransE.}
\label{fig:model}
\end{figure}

In this paper, we aim at extending TransE to model relation paths for representation learning of KBs, and propose path-based TransE (PTransE).  In PTransE, in addition to direct connected relational facts, we also build triples from KBs using entity pairs connected with relation paths. As shown in Figure \ref{fig:model}, TransE only considers direct relations between entities, e.g., $h \xrightarrow{r} t$, builds a triple $(h, r, t)$, and optimizes the objective $\mathbf{h} + \mathbf{r} = \mathbf{t}$. PTransE generalizes TransE by regarding multiple-step relation paths as connections between entities. Take the 2-step path $h \xrightarrow{r_1} e_1 \xrightarrow{r_2} t$ for example as shown in Figure \ref{fig:model}. Besides building triples $(h, r_1, e_1)$ and $(e_1, r_2, t)$ for learning as in TransE, PTransE also builds a triple $(h, r_1 \circ r_2, t)$, and optimizes the objective $\mathbf{h} + (\mathbf{r}_1 \circ \mathbf{r}_2) = \mathbf{t}$, where $\circ$ is an operation to join the relations $r_1$ and $r_2$ together into a unified relation path representation.

As compared with TransE, PTransE takes rich relation paths in KBs for learning. There are two critical challenges that make PTransE nontrivial to learn from relation paths:

\textbf{Relation Path Reliability}. Not all relation paths are meaningful and reliable for learning. For example, there is often a relation path $h \xrightarrow{\texttt{Friend}} e_1 \xrightarrow{\texttt{Profession}} t$, but actually it does not indicate any semantic relationship between $h$ and $t$. Hence, it is inappropriate to consider all relation paths in our model. In experiments, we find that those relation paths that lead to lots of possible tail entities are mostly unreliable for the entity pair. In this paper, we propose a path-constraint resource allocation algorithm to measure the reliability of relation paths. Afterwards, we select the reliable relation paths for representation learning.

\textbf{Relation Path Representation}. In order to take relation paths into consideration, relation paths should also be represented in a low-dimensional space. It is straightforward that the semantic meaning of a relation path depends on all relations in this path. Given a relation path $p = (r_1, \ldots, r_l)$ , 
 we will define and learn a binary operation function ($\circ$) to obtain the path embedding $\mathbf{p}$ by recursively composing multiple relations, i.e., $\mathbf{p} = \mathbf{r}_1 \circ \ldots \circ \mathbf{r}_l$.

With relation path selection and representation, PTransE learns entity and relation embeddings by regarding relation paths as translations between the corresponding entities. In experiments, we select a typical KB, Freebase, to build datasets and carry out evaluation on three tasks, including entity prediction, relation prediction and relation extraction from text. Experimental results show that, PTransE significantly outperforms TransE and other baseline methods on all three tasks.

\section{Our Model}
In this section, we introduce path-based TransE (PTransE) that learns representations of entities and relations considering relation paths. In TransE and PTransE, we have entity set $\mathbf{E}$ and relation set $\mathbf{R}$, and learn to encode both entities and relations in $\mathbb{R}^k$. Given a KB represented by a set of triples $\mathbf{S} = \{(h,r,t)\}$ with each triple composed of two entities $h, t \in \mathbf{E}$ and their relation $r \in \mathbf{R}$. Our model is expected to return a low energy score when the relation holds, and a high one otherwise.

\subsection{TransE and PTransE}
\label{sec:ptranse}
For each triple $(h, r, t)$, TransE regards the relation as a translation vector $\mathbf{r}$ between two entity vectors $\mathbf{h}$ and $\mathbf{t}$. The energy function is defined as
\begin{equation}
\label{eq:transe}
E(h, r, t) = ||\mathbf{h}+\mathbf{r}-\mathbf{t}||,
\end{equation}
which is expected to get a low score when $(h, r, t)$ holds, and high otherwise.

TransE only learns from direct relations between entities but ignores multiple-step relation paths, which also contain rich inference patterns between entities. PTransE take relation paths into consideration for representation learning.

Suppose there are multiple relation paths $P(h, t) = \{p_1, \ldots, p_{N}\}$ connecting two entities $h$ and $t$, where relation path $p = (r_1, \ldots, r_l)$ indicates $h \xrightarrow{r_1} \ldots \xrightarrow{r_l} t$. For each triple $(h, r, t)$, the energy function is defined as
\begin{equation}
\label{eq:ptranse}
G(h, r, t) =  E(h, r, t) + E(h, P, t),
\end{equation}
where $E(h, r, t)$ models correlations between relations and entities with direct relation triples, as defined in Equation (\ref{eq:transe}). $E(h, P, t)$ models the inference correlations between relations with multiple-step relation path triples, which is defined as
\begin{equation}
E(h, P, t) = \frac{1}{Z} \sum_{p \in P(h, t)} R(p|h, t) E(h, p, t),
\end{equation}
where $R(p|h, t)$ indicates the reliability of the relation path $p$ given the entity pair $(h, t)$, $Z = \sum_{p \in P(h, t)}{R(p|h, t)}$ is a normalization factor, and $E(h, p, t)$ is the energy function of the triple $(h, p, t)$.

For the energy of each triple $(h, p, t)$, the component $R(p|h, t)$ concerns about relation path reliability, and $E(h, p, t)$ concerns about relation path representation. We introduce the two components in detail as follows.

\subsection{Relation Path Reliability}
We propose a path-constraint resource allocation (PCRA) algorithm to measure the reliability of a relation path. Resource allocation over networks was originally proposed for personalized recommendation \cite{zhou2007bipartite}, and has been successfully used in information retrieval for measuring relatedness between two objects \cite{lu2011link}. Here we extend it to PCRA to measure the reliability of relation paths. The basic idea is, we assume that a certain amount of resource is associated with the head entity $h$, and will flow following the given path $p$. We use the resource amount that eventually flows to the tail entity $t$ to measure the reliability of the path $p$ as a meaningful connection between $h$ and $t$.

Formally, for a path triple $(h, p, t)$, we compute the resource amount flowing from $h$ to $t$ given the path $p = (r_1, \ldots, r_l)$ as follows. Starting from $h$ and following the relation path $p$, we can write the flowing path as $S_0 \xrightarrow{r_1} S_1 \xrightarrow{r_2} \ldots \xrightarrow{r_l}S_l$, where $S_0 = {h}$ and $t \in S_l$.

For any entity $m \in S_i$, we denote its direct predecessors along relation $r_{i}$ in $S_{i-1}$ as $S_{i-1}(\cdot, m)$. The resource flowing to $m$ is defined as
\begin{equation}
R_p(m) = \sum_{n \in S_{i-1}(\cdot, m)} \frac{1}{|S_{i}(n,\cdot)|} R_p(n),
\end{equation}
where $S_{i}(n,\cdot)$ is the direct successors of $n \in S_{i-1}$ following the relation $r_{i}$, and $R_p(n)$ is the resource obtained from the entity $n$. 

For each relation path $p$, we set the initial resource in $h$ as $R_p(h) = 1$. By performing resource allocation recursively from $h$ through the path $p$, the tail entity $t$ eventually obtains the resource $R_p(t)$ which indicates how much information of the head entity $h$ can be well translated. We use $R_p(t)$ to measure the reliability of the path $p$ given $(h, t)$, i.e., $R(p | h, t) = R_p(t)$.

\subsection{Relation Path Representation}
Besides relation path reliability, we also need to define energy function $E(h, p, t)$ for the path triple $(h, p, t)$ in Equation (\ref{eq:ptranse}). Similar with the energy function of TransE in Equation (\ref{eq:transe}), we will also represent the relation path $p$ in the embedding space. 

\begin{figure}[htb]
\centering
\includegraphics[width=\columnwidth]{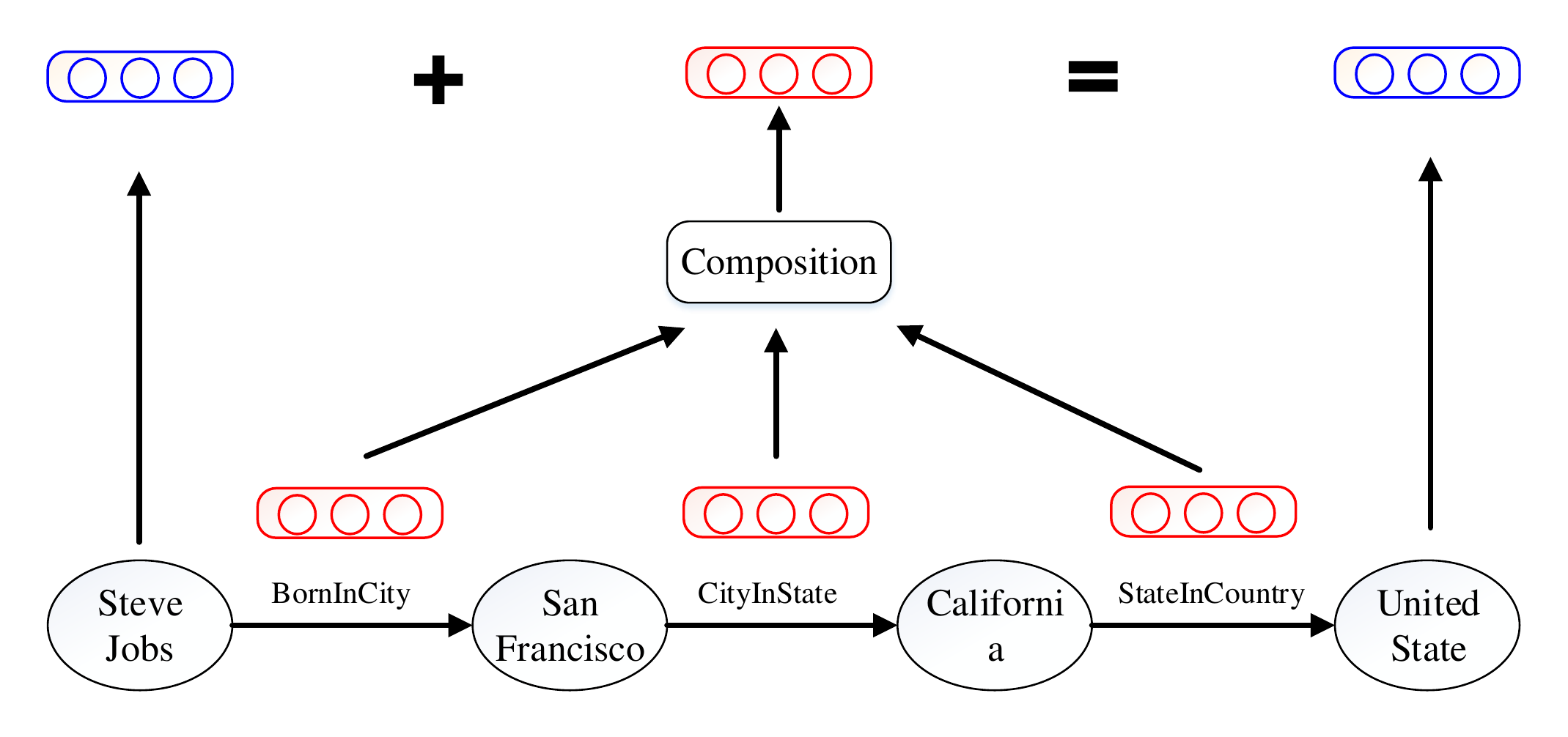}
\caption{Path representations  are computed by semantic composition of relation embeddings.}
\label{fig:path_representation}
\end{figure}

The semantic meaning of a relation path considerably relies on its involved relations. It is thus reasonable for us to build path embeddings via semantic composition of relation embeddings. As illustrated in Figure \ref{fig:path_representation}, the path embedding $\mathbf{p}$ is  composed by the embeddings of \texttt{BorninCity}, \texttt{CityInState} and \texttt{StateInCountry}.

Formally, for a path $p = (r_1, \ldots, r_l)$, we define a composition operation $\circ$ and obtain path embedding as $\mathbf{p} = \mathbf{r}_1 \circ \ldots \circ \mathbf{r}_l$. In this paper, we consider three types of composition operation:

\textbf{Addition (ADD).} The addition operation obtains the vector of a path by summing up the vectors of all relations, which is formalized as
\begin{equation}
\mathbf{p} = \mathbf{r}_1 + \ldots + \mathbf{r}_l.
\end{equation}

\textbf{Multiplication (MUL).} The multiplication operation obtains the vector of a path as the cumulative product of the vectors of all relations, which is formalized as
\begin{equation}
\mathbf{p} = \mathbf{r}_1 \cdot \ldots \cdot \mathbf{r}_l.
\end{equation}
Both addition and multiplication operations are simple and have been extensively investigated in semantic composition of phrases and sentences \cite{mitchell2008vector}.

\textbf{Recurrent Neural Network (RNN).} RNN is a recent neural-based model for semantic composition \cite{mikolov2010recurrent}. The composition operation is realized using a matrix W:
\begin{equation}
c_i = f(W[c_{i-1}; r_i]),
\end{equation}
where $f$ is a non-linearity or identical function, and $[a; b]$ represents the concatenation of two vectors. By setting $c_1 = r_1$ and recursively performing RNN following the relation path, we will finally obtain $p = c_n$. RNN has also been used for representation learning of relation paths in KBs \cite{neelakantan2015compositional}.

For a multiple-step relation path triple $(h, p, t)$, we could have followed TransE and define the energy function as $E(h, p, t) = ||\mathbf{h}+\mathbf{p}-\mathbf{t}||$. However, since we have minimized $||\mathbf{h}+\mathbf{r}-\mathbf{t}||$ with the direct relation triple $(h, r, t)$ to make sure $\mathbf{r} \approx \mathbf{t}-\mathbf{h}$, we may directly define the energy function of $(h, p, t)$ as
\begin{equation}
E(h, p, t) = ||\mathbf{p}-(\mathbf{t} - \mathbf{h})|| = ||\mathbf{p}-\mathbf{r}|| = E(p, r),
\end{equation}
which is expected to be a low score when the multiple-relation path $p$ is consistent with the direct relation $r$, and high otherwise, without using entity embeddings.


\subsection{Objective Formalization}
We formalize the optimization objective of PTransE as
\begin{equation}
\label{loss-function}\small
L(\mathbf{S}) = \sum_{(h, r, t) \in \mathbf{S}} \big[L(h, r, t) + \frac{1}{Z} \sum_{p \in P(h, t)} R(p|h, t) L(p, r)\big].
\end{equation}
Following TransE, $L(h, r, t )$ and $L(p, r)$ are margin-based loss functions with respect to the triple $(h, r, t)$ and the pair $(p, r)$:
\begin{equation}
\small
L(h, r, t) = \sum_{(h', r', t') \in \mathbf{S}^{-}} [ \gamma + E(h, r, t)- E(h', r', t')]_+,
\end{equation}
and
\begin{equation}
\small
L(p, r) = \sum_{(h, r', t) \in \mathbf{S}^{-}} [ \gamma + E(p, r)- E(p, r')]_+,
\end{equation}
where $[x]_+ = \max(0,x)$ returns the maximum between $0$ and $x$, $\gamma$ is the margin, $\mathbf{S}$ is the set of valid triples existing in a KB and $\mathbf{S}^{-}$ is the set of invalid triples. The objective will favor lower scores for valid triples as compared with invalid triples.

The invalid triple set with respect to $(h, r, t)$ is defined as
\begin{equation}
\mathbf{S}^{-} = \{(h', r, t)\} \cup  \{(h, r', t)\} \cup  \{(h, r, t')\}.
\end{equation}
That is, the set of invalid triples is composed of the original valid triple $(h, r, t)$ with one of three components replaced.

\subsection{Optimization and Implementation Details}
For optimization, we employ stochastic gradient descent (SGD) to minimize the loss function. We randomly select a valid triple from the training set iteratively for learning. In the implementation, we also enforce constraints on the norms of the embeddings $h$, $r$, $t$. That is, we set
\begin{equation}
\|\mathbf{h}\|_2\le 1, \quad \|\mathbf{r}\|_2\le 1, \quad \|\mathbf{t}\|_2\le1. \quad \forall h, r, t.
\end{equation}

There are also some implementation details that will significantly influence the performance of representation learning, which are introduced as follows.

\textbf{Reverse Relation Addition.} In some cases, we are interested in the reverse version of a relation, which may not be presented in KBs. For example, according to the relation path $e_1 \xrightarrow{\texttt{BornInCity}} e_2 \xleftarrow{\texttt{CityOfCountry}} e_3$ we expect to infer the fact that $(e_1, \texttt{Nationality}, e3)$. In this paper, however, we only consider the relation paths following one direction. Hence, we add reverse relations for each relation in KBs. That is, for each triple $(h, r, t)$ we build another $(t, r^{-1}, h)$. In this way, our method can consider the above-mentioned path as $e_1 \xrightarrow{\texttt{BornInCity}} e_2 \xrightarrow{\texttt{CityOfCountry}^{-1}} e_3$ for learning.

\textbf{Path Selection Limitation.} There are usually large amount of relations and facts about each entity pair. It will be impractical to enumerate all possible relation paths between head and tail entities. For example, if each entity refers to more than $100$ relations on average, which is common in Freebase, there will be billions of $4$-step paths. Even for $2$-step or $3$-step paths, it will be time-consuming to consider all of them without limitation. For computational efficiency, in this paper we restrict the length of paths to at most $3$-steps and consider those relation paths with the reliability score larger than $0.01$.

\subsection{Complexity Analysis}
We denote $N_e$ as the number of entities, $N_r$ as the number of relations and $K$ as the vector dimension. The model parameter size of PTransE is $(N_e K + N_r K)$, which is the same as TransE. PTransE follows the optimization procedure introduced by \cite{bordes2013translating} to solve Equation (\ref{loss-function}). We denote $S$ as the number of triples for learning, $P$ as the expected number of relation paths between two entities, and $L$ as the expected length of relation paths. For each iteration in optimization, the complexity of TransE is $O(SK)$ and the complexity of PTransE is $O(SKPL)$ for ADD and MUL, and $O(SK^2PL)$ for RNN.

\section{Experiments and Analysis}

\subsection{Data Sets and Experimental Setting}
We evaluate our method on a typical large-scale KB Freebase \cite{bollacker2008freebase}. In this paper, we adopt two datasets extracted from Freebase, i.e., FB15K and FB40K. The statistics of the datasets are listed in Table \ref{table:statistics}.

\begin{table}[htb]
\centering \small
\caption{Statistics of data sets.}
\label{table:statistics}
\begin{tabular}{|c|rrrrr|}
\hline
Dataset &\#Rel& \#Ent& \#Train& \#Valid& \# Test\\
\hline
FB15K & 1,345 & 14,951 & 483,142 & 50,000 & 59,071 \\
FB40K & 1,336 & 39,528 & 370,648 & 67,946 & 96,678 \\
\hline
\end{tabular}
\end{table}

We evaluate the performance of PTransE and other baselines by predicting whether testing triples hold. We consider two scenarios: (1) Knowledge base completion, aiming to predict the missing entities or relations in given triples only based on existing KBs. (2) Relation extraction from texts, aiming to extract relations between entities based on information from both plain texts and KBs.

\subsection{Knowledge Base Completion}
The task of knowledge base completion is to complete the triple $(h,r,t)$ when one of $h, t, r$ is missing. The task has been used for evaluation in \cite{bordes2011learning,bordes2012joint,bordes2013translating}. We conduct the evaluation on FB15K, which has $483,142$ relational triples and $1,345$ relation types, among which there are rich inference and reasoning patterns.

In the testing phase, for each testing triple $(h, r, t)$, we define the following score function for prediction
\begin{equation}
S(h, r, t) = G(h, r, t) + G(t, r^{-1}, h),
\end{equation}
and the score function $G(h, r, t)$ is further defined as
\begin{align}
G(h, r, t) =& ||\mathbf{h} + \mathbf{r} - \mathbf{t}|| + \nonumber \\
&\frac{1}{Z} \sum_{p\in P(h,t)}\Pr(r|p)R(p|h, t)||\mathbf{p}-\mathbf{r}||.
\end{align}
The score function is similar to the energy function defined in Section \ref{sec:ptranse}. The difference is that, here we consider the  reliability of a path $p$ is also related to the inference strength given $r$, which is quantified as $\Pr(r|p) = \Pr(r,p) / \Pr(p)$ obtained from the training data.

We divide the stage into two sub-tasks, i.e., entity prediction and relation prediction.

\subsubsection{Entity Prediction}
In the sub-task of entity prediction, we follow the setting in \cite{bordes2013translating}. For each testing triple with missing head or tail entity, various methods are asked to compute the scores of $G(h, r, t)$ for all candidate entities and rank them in descending order.

We use two measures as our evaluation metrics: the mean of correct entity ranks and the proportion of  valid entities ranked in top-$10$ (Hits@10). As mentioned in \cite{bordes2013translating}, the measures are desirable but flawed when an invalid triple ends up being valid in KBs. For example, when the testing triple is (\emph{Obama}, \texttt{PresidentOf}, \emph{USA}) with the head entity \emph{Obama} is missing, the head entity \emph{Lincoln} may be regarded invalid for prediction, but in fact it is valid in KBs. The evaluation metrics will under-estimate those methods that rank these triples high. Hence, we can filter out all these valid triples in KBs before ranking. The first evaluation setting was named as ``Raw'' and the second one as ``Filter''.

For comparison, we select all methods in \cite{bordes2013translating,wang2014knowledge} as our baselines and use their reported results directly since the evaluation dataset is identical.

Ideally, PTransE has to find all possible relation paths between the given entity and each candidate entity. However, it is time consuming and impractical, because we need to iterate all candidate entities in $|E|$ for each testing triple. Here we adopt a re-ranking method: we first rank all candidate entities according to the scores from TransE, and then re-rank top-$500$ candidates according to the scores from PTransE.

For PTransE, we find the best hyperparameters according to the mean rank in validation set. The optimal configurations of PTransE we used are $\lambda = 0.001$, $\gamma = 1$, $k = 100$ and taking $L_1$ as dissimilarity. For training, we limit  the number of epochs   over all the training triples to  $500$.

\begin{table}[htb]
\centering \small
\caption{Evaluation results on entity prediction.}
\label{label:link_prediction_entity}
\begin{tabular}{|c|rr|rr|}
\hline
\multirow{2}{*}{Metric}& \multicolumn{2}{|c|}{Mean Rank}& \multicolumn{2}{|c|}{Hits@10 ($\%$)} \\
& Raw & Filter & Raw & Filter \\
\hline
RESCAL        &    828 & 683 &  28.4 & 44.1\\
SE            &    273 & 162 &  28.8 & 39.8\\
SME (linear)  &    274 & 154 &  30.7 & 40.8\\
SME (bilinear)&    284 & 158 &  31.3 & 41.3\\
LFM           &    283 & 164 &  26.0 & 33.1\\
TransE        &    243 & 125 &  34.9 & 47.1\\
TransH        &    212 &  87 &  45.7 & 64.4\\
TransR        &    \textbf{198}& 77 &  48.2 & 68.7 \\
TransE (Our)  &    205 &  63 &  47.9 & 70.2 \\
\hline
PTransE (ADD, 2-step) &    200 & \textbf{54}& \textbf{51.8} & 83.4\\
PTransE (MUL, 2-step) &    216 &  67 & 47.4& 77.7 \\
PTransE (RNN, 2-step) & 242 & 92 & 50.6 & 82.2 \\
PTransE (ADD, 3-step) &207 & 58 & 51.4 & \textbf{84.6} \\
\hline
\end{tabular}
\end{table}

Evaluation results of entity prediction are shown in Table \ref{label:link_prediction_entity}. The baselines include RESCAL \cite{nickel2011three}, SE \cite{bordes2011learning}, SME (linear) \cite{bordes2012joint}, SME (bilinear) \cite{bordes2012joint}, LFM  \cite{jenatton2012latent}, TransE \cite{bordes2013translating} (original version and our implementation considering reverse relations), TransH \cite{wang2014knowledge}, and TransR \cite{lin2015learning}.

For PTransE, we consider three composition operations for relation path representation: addition (ADD), multiplication (MUL) and recurrent neural networks (RNN). We also consider relation paths with at most 2-steps and 3-steps. With the same configurations of PTransE, our TransE implementation achieves much better performance than that reported in \cite{bordes2013translating}.

From Table \ref{label:link_prediction_entity} we observe that: (1) PTransE significantly and consistently outperforms other baselines including TransE. It indicates that relation paths provide a good supplement for representation learning of KBs, which have been successfully encoded by PTransE. For example, since both \emph{George W. Bush} and \emph{Abraham Lincoln} were presidents of \emph{the United States}, they exhibit similar embeddings in TransE. This may lead to confusion for TransE to predict the spouse of \emph{Laura Bush}. In contrast, since PTransE models relation paths, it can take advantage of the relation paths between \emph{George W. Bush} and \emph{Laura Bush}, and leads to more accurate prediction. (2) For PTransE, the addition operation outperforms other composition operations in both Mean Rank and Hits@10. The reason is that, the addition operation is compatible with the learning objectives of both TransE and PTransE. Take $h \xrightarrow{r_1} e_1 \xrightarrow{r_2} t$ for example. The optimization objectives of two direct relations $\mathbf{h} + \mathbf{r}_1 = \mathbf{e}_1$ and $\mathbf{e}_1 + \mathbf{r}_2 = \mathbf{t}$ can easily derive the path objective $\mathbf{h} + \mathbf{r}_1 + \mathbf{r}_2 = \mathbf{t}$. (3) PTransE of considering relation paths with at most 2-step and 3-step achieve comparable results. This indicates that it may be unnecessary to consider those relation paths that are too long.

\begin{table*}[htb]
\centering
\caption{Evaluation results on FB15K by mapping properties of relations. ($\%$)}
 \label{label:mapping_property}
\begin{tabular}{|c|rrrr|rrrr|}
\hline
Tasks &\multicolumn{4}{|c|}{Predicting Head Entities (Hits@10)}&\multicolumn{4}{|c|}{Predicting Tail Entities (Hits@10)}\\
\hline
Relation Category & 1-to-1 & 1-to-N & N-to-1 & N-to-N & 1-to-1 & 1-to-N & N-to-1 & N-to-N \\
\hline
SE                   & 35.6  & 62.6 & 17.2 & 37.5 & 34.9 & 14.6 & 68.3 & 41.3\\
SME (linear)         & 35.1  & 53.7 & 19.0 & 40.3 & 32.7 & 14.9 & 61.6 & 43.3\\
SME (bilinear)       & 30.9  & 69.6 & 19.9 & 38.6 & 28.2 & 13.1 & 76.0 & 41.8\\
TransE               & 43.7  & 65.7 & 18.2 & 47.2 & 43.7 & 19.7 & 66.7 & 50.0\\
TransH               & 66.8  & 87.6 &  28.7 & 64.5 & 65.5 & 39.8 & 83.3 & 67.2\\
TransR               & 78.8  &89.2&34.1&69.2&79.2&37.4&\textbf{90.4}&72.1 \\
TransE (Our) &74.6& 86.6& 43.7& 70.6&71.5& 49.0& 85.0&72.9\\
\hline
PTransE (ADD, 2-step) &\textbf{91.0} &\textbf{92.8} &\textbf{60.9} &83.8&\textbf{91.2} &\textbf{74.0} &88.9 &86.4\\
PTransE (MUL, 2-step) &89.0&86.8&57.6&79.8&87.8&71.4&72.2&80.4\\
PTransE (RNN, 2-step) & 88.9 &84.0 &56.3 & 84.5 & 88.8 & 68.4 & 81.5 &86.7\\
PTrasnE (ADD, 3-step) & 90.1 &92.0 & 58.7& \textbf{86.1} & 90.7 & 70.7 & 87.5 & \textbf{88.7} \\
\hline
\end{tabular}
\end{table*}

As defined in \cite{bordes2013translating}, relations in KBs can be divided into various types according to their mapping properties such as 1-to-1, 1-to-N, N-to-1 and N-to-N. Here we demonstrate the performance of PTransE and some baselines with respect to different types of relations in Table \ref{label:mapping_property}. It is observed that, on all mapping types of relations, PTransE consistently achieves significant improvement as compared with TransE.

\subsubsection{Relation Prediction}
Relation prediction aims to predict relations given two entities. We also use FB15K for evaluation. In this sub-task, we can use the score function of PTransE to rank candidate relations instead of re-ranking like in entity prediction.

Since our implementation of TransE has achieved the best performance among all baselines for entity prediction, here we only compare PTransE with TransE due to limited space. Evaluation results are shown in Table \ref{label:link_prediction_rel}, where we report Hits@1 instead of Hits@10 for comparison, because Hits@10 for both TransE and PTransE exceeds 95\%. In this table, we report the performance of TransE without reverse relations (TransE), with reverse relations (+Rev) and considering relation paths for testing like that in PTransE (+Rev+Path). We report the performance of PTransE with only considering relation paths (-TransE),  only considering the part in Equation (\ref{eq:transe}) (-Path) and  considering both (PTransE).

The optimal configurations of PTransE for relation prediction are identical to those for entity prediction: $\lambda = 0.001$, $\gamma = 1$, $k = 100$ and taking $L_1$ as dissimilarity.

From Table \ref{label:link_prediction_rel} we observe that: (1) PTransE outperforms TransE+Rev+Path significantly for relation prediction by reducing $41.8\%$ prediction errors. (2) Even for TransE itself, considering relation paths for testing can reduce $17.3\%$ errors as compared with TransE+Rev. It indicates that encoding relation paths will benefit for predicting relations. (3) PTransE with only considering relation paths (PTransE-TransE) gets surprisingly high mean rank. The reason is that, not all entity pairs in testing triples have relation paths, which will lead to random guess and the expectation of rank of these entity pairs is $|R|/2$. Meanwhile, Hits@1 of PTransE-TransE is relatively reasonable, which indicates the worth of modeling relation paths. As compared with TransE, the inferior of PTransE-TransE also indicates that entity representations are informative and crucial for relation prediction.

\begin{table}[htb]
\centering \small
\caption{Evaluation results on relation prediction.}
\label{label:link_prediction_rel}
\begin{tabular}{|l|rr|rr|}
\hline
\multirow{2}{*}{Metric} & \multicolumn{2}{|c|}{Mean Rank} & \multicolumn{2}{|c|}{Hits@1 ($\%$)} \\
           & Raw &Filter & Raw & Filter \\ \hline
TransE (Our)      & 2.8 & 2.5 & 65.1 & 84.3\\
$\quad$+Rev       & 2.6 & 2.3 & 67.1 & 86.7\\
$\quad$+Rev+Path  & 2.4 & 1.9 & 65.2 & 89.0\\ \hline
PTransE (ADD, 2-step) & \textbf{1.7}&    \textbf{1.2}& 69.5& 93.6\\
$\quad$-TransE  & 135.8&135.3&51.4&78.0 \\
$\quad$-Path    & 2.0 & 1.6 & \textbf{69.7}& 89.0\\
PTransE (MUL, 2-step) & 2.5 & 2.0 & 66.3 & 89.0\\
PTransE (RNN, 2-step) & 1.9 &1.4&68.3&93.2\\
PTransE (ADD, 3-step) & 1.8 & 1.4 &68.5 & \textbf{94.0} \\
\hline
\end{tabular}
\end{table}

\subsection{Relation Extraction from Text}
Relation extraction from text aims to extract relational facts from plain text to enrich existing KBs. Many works regard large-scale KBs as distant supervision to annotate sentences as training instances and build relation classifiers according to features extracted from the sentences \cite{mintz2009distant,riedel2010modeling,hoffmann2011knowledge,surdeanu2012multi}. All these methods reason new facts only based on plain text. TransE was used to enrich a text-based model and achieved a significant improvement \cite{weston2013connecting}, and so do TransH \cite{wang2014knowledge} and TransR \cite{lin2015learning}. In this task, we explore the effectiveness of PTransE for relation extraction from text.

We use New York Times corpus (NYT)  released by \cite{riedel2010modeling} as training and testing data. NYT aligns Freebase with the articles in New York Times, and extracts sentence-level features such as part-of-speech tags, dependency tree paths for each mention. There are $53$ relations (including non-relation denoted as \texttt{NA}) and $121,034$ training mentions. We use FB40K as the KB, consisting all entities mentioned in NYT and $1,336$ relations.

\begin{figure}[htb]
\centering
\includegraphics[width=\columnwidth]{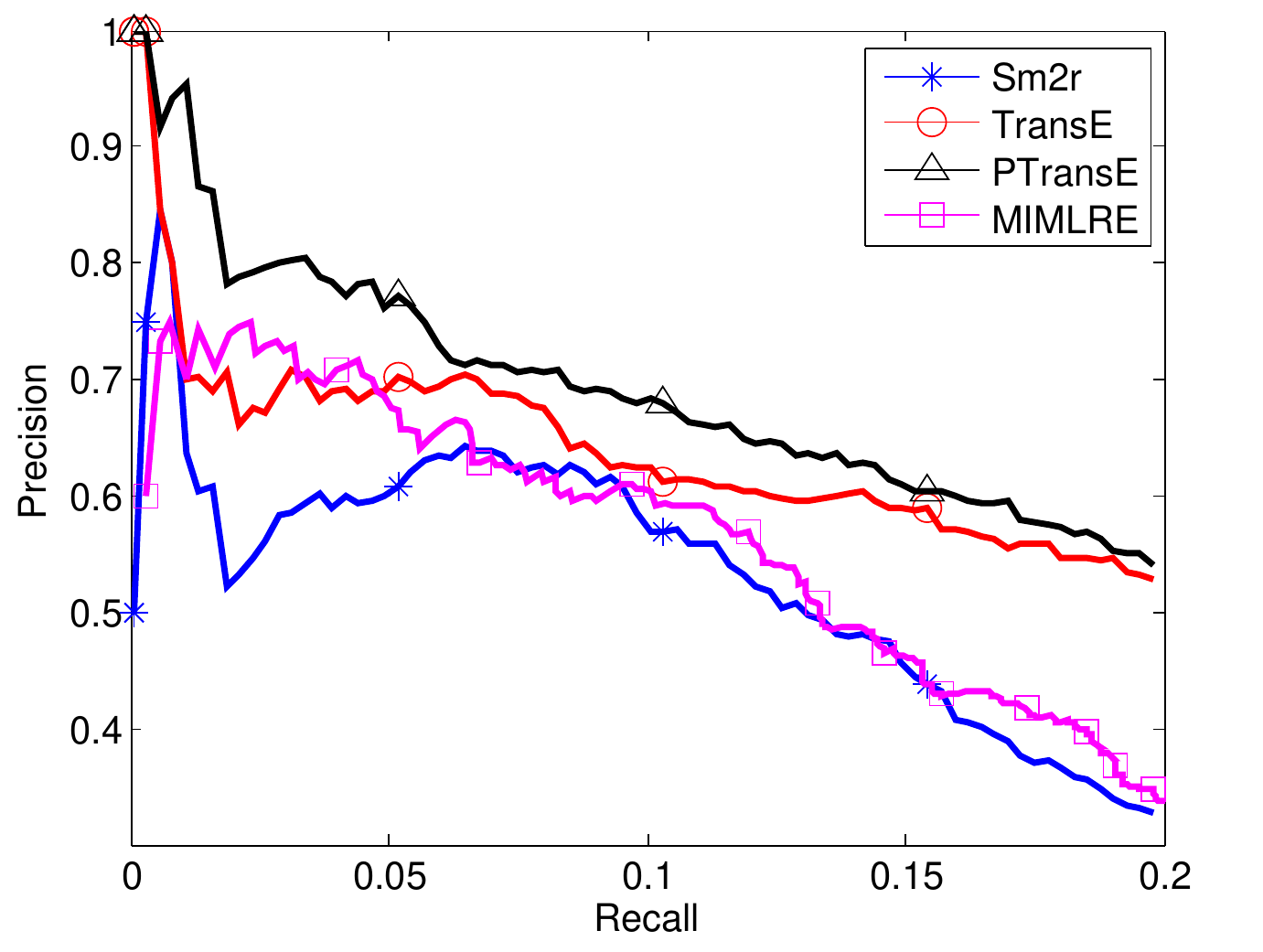}
\caption{Precision-recall curves of Sm2r, TransE and PTransE combine with Sm2r.}
\label{fig:relation_extraction}
\end{figure}

In the experiments, we implemented the text-based model Sm2r presented in \cite{weston2013connecting}. We combine the ranking scores from the text-based model with those from KB representations to rank testing triples, and generate precision-recall curves for both TransE and PTransE. For learning of TransE and PTransE, we set the dimensions of entities/relations embeddings $k = 50$, the learning rate $\lambda= 0.001$, the margin $\gamma = 1.0$ and dissimilarity metric as L1. We also compare with MIMLRE \cite{surdeanu2012multi} which is the state-of-art method using distant supervision. The evaluation curves are shown in Figure \ref{fig:relation_extraction}.

From Figure \ref{fig:relation_extraction} we can observe that, by combining with the text-based model Sm2r, the precision of PTransE significantly outperforms TransE especially for the top-ranked triples. This indicates that encoding relation paths is also useful for relation extraction from text.

Note that TransE used here does not consider reverse relations and relation paths because the performance does not change much. We analyze the reason as follows. In the task of knowledge base completion, each testing triple has at least one valid relation. In contrast, many testing triples in this task correspond to non-relation (\texttt{NA}), and there are usually several relation paths between two entities in these non-relation triples. TransE does not encode relation paths during the training phase like PTransE, which results in worse performance for predicting non-relation when considering relation paths in the testing phase, and compensates the improvement on those triples that do have relations. This indicates it is non-trivial to encode relation paths, and also confirms the effectiveness of PTransE.

\subsection{Case Study of Relation Inference}

We have shown that PTransE can achieve high performance for knowledge base completion and relation extraction from text. In this section, we present some examples of relation inference according to relation paths.

\begin{figure}[htb]
\centering
\includegraphics[width=\columnwidth]{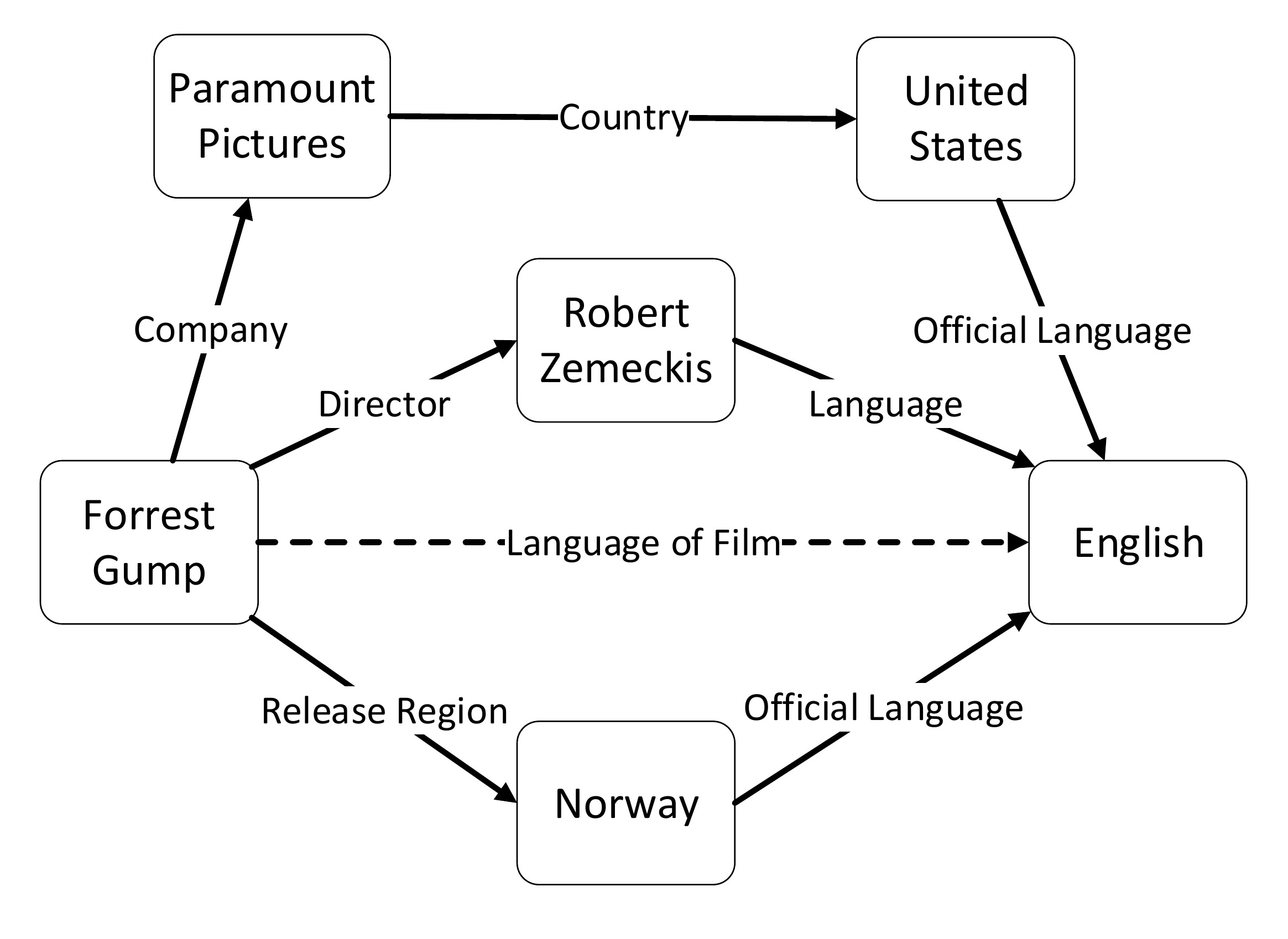}
\caption{An inference example in Freebase.}
\label{fig:path}
\end{figure}

According to the learning results of PTransE, we can find new facts from KBs. As shown in Figure \ref{fig:path}, two entities \emph{Forrest Gump} and \emph{English} are connected by three relation paths, which give us more confidence to predict the relation between the two entities to \texttt{LanguageOfFilm}.

\section{Related Work}
Recent years have witnessed great advances of modeling multi-relational data such as social networks and KBs. Many works cope with relational learning as a multi-relational representation learning problem, encoding both entities and relations in a low-dimensional latent space, based on Bayesian clustering \cite{kemp2006learning,miller2009nonparametric,sutskever2009modelling,zhu2012max}, energy-based models \cite{bordes2011learning,chen2013learning,socher2013reasoning,bordes2013translating,bordes2014semantic}, matrix factorization \cite{singh2008relational,nickel2011three,nickel2012factorizing} . Among existing representation models, TransE \cite{bordes2013translating} regards a relation as translation between head and tail entities for optimization, which achieves a good trade-off between prediction accuracy and computational efficiency. All existing representation learning methods of knowledge bases only use direct relations between entities, ignoring rich information in relation paths.

Relation paths have already been widely considered in social networks and recommender systems. Most of these works regard each relation and path as discrete symbols, and deal with them using graph-based algorithms, such as random walks with restart \cite{tong2006fast}. Relation paths have also been used for inference on large-scale KBs, such as Path Ranking algorithm (PRA) \cite{lao2010relational}, which has been adopted for expert finding \cite{lao2010relational} and information retrieval \cite{lao2012reading}. PRA has also been used for relation extraction based on KB structure \cite{lao2011random,gardner2013improving}. \cite{neelakantan2015compositional} further learns a recurrent neural network (RNN) to represent unseen relation paths according to involved relations. We note that, these methods focus on modeling relation paths for relation extraction without considering any information of entities. In contrast, by successfully integrating the merits of modeling entities and relation paths, PTransE can learn superior representations of both entities and relations for knowledge graph completion and relation extraction as shown in our experiments.

\section{Conclusion and Future Work}
This paper presents PTransE, a novel representation learning method for KBs, which encodes relation paths to embed both entities and relations in a low-dimensional space. To take advantages of relation paths, we propose path-constraint resource allocation to measure relation path reliability, and employ semantic composition of relations to represent paths for optimization. We evaluate PTransE on knowledge base completion and relation extraction from text. Experimental results show that PTransE achieves consistent and significant improvements as compared with TransE and other baselines.

In future, we will explore the following research directions: (1) This paper only considers the inference patterns between direct relations and relation paths between two entities for learning. There are much complicated patterns among relations. For example, the inference form $\texttt{Queen}(e) \xRightarrow{\text{Inference}} \texttt{Female}(e)$ cannot be handled by PTransE. We may take advantages of first-order logic to encode these inference patterns for representation learning. (2) There are some extensions for TransE, e.g., TransH and TransR. It is non-trivial for them to adopt the idea of PTransE, and we will explore to extend PTransE to these models to better deal with complicated scenarios of KBs.

\section{ Acknowledgments}
Zhiyuan Liu and Maosong Sun are supported by the 973 Program (No. 2014CB340501) and the National Natural Science Foundation of China (NSFC No. 61133012) and Tsinghua-Samsung Joint Lab. Huanbo Luan is supported by the National Natural Science Foundation of China (NSFC No. 61303075). We sincerely thank Yansong Feng for insightful discussions, and thank all anonymous reviewers for their constructive comments.

\bibliographystyle{acl}
\bibliography{relation_path}
\end{document}